\documentclass{article}
\usepackage[utf8]{inputenc}

\usepackage[numbers]{natbib}
\usepackage{graphicx}
\usepackage{authblk}
\usepackage{booktabs}
\usepackage{multirow,multicol}
\usepackage{xcolor}
\usepackage{colortbl}
\usepackage{collcell}
\usepackage{hyperref}
\usepackage{amsfonts}
\usepackage{enumitem}

\newcommand{\x}{{\bf x}}
\newcommand{\z}{{\bf z}}
\newcommand{\bb}[1]{\mathbb{#1}}

\providecommand{\keywords}[1]
{
	\small	
	\textbf{\textit{Keywords---}} #1
}

\title{Feature Selection from High-Dimensional Data with Very Low Sample Size: A Cautionary Tale}

\author[1]{Ludmila~I. Kuncheva}
\author[1]{Clare~E. Matthews}
\author[2]{\'Alvar Arnaiz-Gonz\'alez}
\author[2]{Juan~J.~Rodr\'iguez}
\affil[1]{School of Computer Science and Electronic Engineering, Bangor University}
\affil[2]{Escuela Polit\'ecnica Superior, Universidad de Burgos}

\usepackage{natbib}
\usepackage{graphicx}

\begin{document}

\maketitle

\begin{abstract}
	In classification problems, the purpose of feature selection is to identify a small, highly discriminative subset of the original feature set. 
	In many applications, the dataset may have thousands of features and only a few dozens of samples (sometimes termed `wide'). This study is a cautionary tale demonstrating why feature selection in such cases may lead to undesirable results. In view to highlight the sample size issue, we derive the required sample size for declaring two features different. Using an example, we illustrate the heavy dependency between feature set and classifier, which poses a question to classifier-agnostic feature selection methods. However, the choice of a good selector-classifier pair is hampered by the low correlation between estimated and true error rate, as illustrated by another example. While previous studies raising similar issues validate their message with mostly synthetic data, here we carried out an experiment with 20 real datasets. We created an exaggerated scenario whereby we cut a very small portion of the data (10 instances per class) for feature selection and used the rest of the data for testing. The results reinforce the caution and suggest that it may be better to refrain from feature selection from very wide datasets rather than return misleading output to the user.
\end{abstract}


\section{Introduction}
\label{sec:introduction}

Feature selection is a long-standing theme in pattern recognition and machine learning, which inspires researchers to this day~\cite{Bolon13,Chandrashekar14,Bolon15,Storcheus15,Miao16,Li16,Li17,Li17a,Cai18,Bommert20}. 
It is a prolific field where new methods and techniques appear continually and are applied to emerging tasks such as multi-label learning~\cite{Spolaor16,Pereira18} and big data~\cite{Bolon15,Ramirez18}.
Recent feature selection approaches draw predominantly from   information-theory~\cite{Vergara14,Brown12}, ensembles~\cite{BolonCanedo19}, evolutionary learning~\cite{Xue16,Hancer18}, and deep-learning~\cite{Zou15}. A bibliographic search on Web-of-Science\footnote{\url{https://wok.mimas.ac.uk/}} reveals that feature selection is a popular research area with nearly 25\,000 publications containing ``feature selection'' in title or abstract in the past 10 years (as of 7th August 2019). Web-of-Science gives a conservative estimate; hence the real number of publications is likely much larger. Alongside the general-purpose feature selection algorithms~\cite{Hancer18,Bennasar15}, domain-specific methods are being developed for fields such as bioinformatics~\cite{BolonCanedo14,Ditzler12}, text categorization~\cite{Deng19}, and multimedia~\cite{Lee17}. Recent surveys and reviews provide much needed guidance of supervised~\cite{Miao16}, semi-supervised~\cite{Sheikhpour17}, and unsupervised~\cite{Solorio19} feature selection.

Present day data is typically characterised by a large number of features, prime examples of which are bioinformatics~\cite{Bolon15,Saeys07} and multimedia~\cite{Li17a,Lee17}. In all this wealth of literature, cautionary tales about the inadequacy of feature selection for very small-sized data are often overlooked. 

Selecting a subset of features from data with a very small number of samples and a large number of features poses a serious challenge~\cite{Bolon13,EinDor06,Fan09,Liu10,Dernoncourt14,Krawczuk16,Varoquax18}. While not universally accepted, it is convenient to term such data sets `wide'~\cite{Tuv09,Izetta17}.
To gauge the scale of growth of the problem, consider a study by Murray in 1977, where a set of 157 features was described as `embarrassingly large'~\cite{Murray77}. 
In today's terms, a wide dataset would have a few dozens of instances and possibly thousands of features. 

The main problem with wide datasets is the possibility of dramatically overfitting the data. Added to this problem is another one, which we have no control over -- the representativeness of the available sample. Overfitting a non-representative sample may return a useless feature set. In the spirit of Occam's razor, the feature selection method will have to be kept to the simplest possible choice. This eliminates a vast and very successful category of feature selection methods, whose main focus is the traversing of the possible feature subsets. This category includes evolutionary algorithms as well other nature-inspired feature selection methods such as swarm optimisation, particle optimisation, ant colony, bees, grey wolf, cuttlefish, bat, and many more~\cite{Xue16,Zawbaa18}. Such algorithms can be very successful for high-dimensional data but only when the sample size is adequate. In many studies, both recent and past, sophisticated algorithms for selecting features out of a set of a few thousands are evaluated on relatively small data sets. While we are considering an extreme case, where each class contains only 10 examples, the concerns we raise here still apply, albeit to a lesser extent.

The success of a feature selection attempt from a wide dataset depends on several factors: the underlying probability distributions (some problems could be easy to solve), the sample size (number of instances), the dimensionality (number of features), the chosen method for feature selection (how well it discovers good feature subsets, how robust it is to overfitting, how accurately the criterion of interest is evaluated), and the classifier subsequently recommended to the user. 
The literature on feature selection from wide datasets identifies the following caveats:

\begin{enumerate}
	\item  {\em Low-quality subset.} More often than not, feature selection algorithms produce feature subsets whose classification error is not close to that of the optimal feature set~\cite{Sima06}, and sometimes is far in excess of it~\cite{Sima05b}. 
	Alternatively, due to the very small sample size, spurious feature subsets could be returned to the user, reporting, at the same time, a deceptively low error rate. 
	
	\item {\em Low correlation between estimated and true error.} The true error does not correlate well with its estimate even for the best error estimators~\cite{BragaNeto04c,Varoquaux18,Dougherty10}. It has been argued that the quality of the estimate of the error criterion is  more important than the feature selection method itself~\cite{Sima05b}.  
	
	\item {\em Inadequate feature selection protocol.} Many studies including recent ones are oblivious to the `peeking' practice whereby feature selection is carried out first using the available data, and then the chosen subset is estimated through some cross-validation protocol {\em on the same data}. The inadequacy of this protocol has been flagged many times over the years~\cite{Saeys07,Reunanen03,Schulerud04,Singhi06,Rafaeilzadeh07,Smialwski10,Diciotti13,Aldehim17,KunchevaPR18}, warning the practitioners about the high optimistic bias of the classification error. The `proper' protocol would include the feature selection followed by the classification in the cross-validation loop, thereby evaluating the error of the whole process.  (The `proper/wrong' protocols~\cite{KunchevaPR18} are called respectively `IN/OUT' by Rafaeilzadeh et al.~\cite{Rafaeilzadeh07} and `PART/ALL' by Aldehim and Wang~\cite{Aldehim17}.) Probably due to its computational burden, the `proper' protocol has not been widely adopted yet.   
\end{enumerate}

The alarming conclusions from these observations are that: 
\begin{itemize}
	\item From 1) and 2): The subset of features returned to the user may not be adequate.
	\item From 3): The classification accuracy/error predicted for the selected subset may be deceptively low.
\end{itemize}

This study is intended as a cautionary tale. Our contributions are summarised as follows:

\begin{itemize}
	\item We show that the necessary sample size for declaring two features significantly different is very large, which is a warning against selecting features from very small-size data.
	\item We exemplify issues 1) and 2) and examine the source of the discrepancy between between training and testing error.
	\item Using a synthetic example, we demonstrate a flaw of the classifier-agnostic feature selection methods.
	\item To demonstrate the grounds for our concern, we carry out an experimental study with 20 real, benchmark datasets. This is in contrast with  previous studies on feature selection from wide data which are mostly carried out on synthetic, well-behaved data~\cite{Dougherty10,Schulerud04,Sima05,BragaNeto04}.
\end{itemize}

The rest of the paper is organised as follows. Most of the related work is summarised in this introduction, and where suitable in the following sections. Section~\ref{rel} explains the problem and the various types of error in feature selection. The effect of sample size is discussed in Section~\ref{samsz}. The experiment is presented in Section~\ref{exp}, and the conclusion, in Section~\ref{con}.

\section{Sample size for declaring two features different}
\label{samsz}

How big a sample do we need for feature selection? An interesting study on the required number of samples needed for returning a reliable set of features was published by Ein-Dor et al.~\cite{EinDor06,EinDor06a}. These studies came from the area of clinical cancer research where the task is to analyse gene expression data and predict the outcome for a given patient, that is, the potential for relapse and for metastasis. The authors were  inspired by the observation that several research groups had published vastly different sets of predictive genes obtained for the same clinical types of 	patients. The reported lists of genes  differed widely and had only very few genes in common. The authors build a statistical model of the degree of intersection of two sets of features selected from different samples. The implied feature selection method is a filter which ranks the features according to the absolute value of their correlation with the (binary) class label variable. 
The authors assume that the correlation $\rho$ between feature $x_i$ and the (binary) class label variable $y$ is a measure of quality of the feature. They proceed to transform this correlation into a new variable $z_i = \tanh^{-1} (\rho(x_i,y))$, which has a normal distribution. In gene expression data, it may be reasonable to assume that only positive correlation is informative. We note, however, that negative and positive correlations are equally valuable in general feature selection. (Should we, however, transform $|\rho|$ instead of $\rho$, we lose the normal distribution property of $z_i$, and the assumptions for the derivations do not hold.) The staggering conclusion of Ein-Dor et al.'s  work~\cite{EinDor06,EinDor06a} is that thousands of instances are needed even for moderate agreement between the lists (disputed recently on the ground of Gaussianity and sparsity of the data~\cite{Jacobovic18}). 

Here we present a similarly discouraging result for just two features.

\subsection{Counting estimator}
An error estimator is a natural alternative to correlation as a measure of a feature quality, although it may suffer from the inconvenience of the discretisation for small values of the sample size $N$. Let $p_i$ be the probability that a classifier built on feature $x_i$ will label correctly a randomly drawn object from the distribution of the problem at hand. Assuming only two outputs (correct/incorrect), we can form a Bernoulli variable $w_i$, whose probability of success is $p_i$. Consider features $x_1$ and $x_2$. The task is to pick the better feature.  Without loss of generality, let $p_1 > p_2$. In this case, we should prefer $x_1$ to $x_2$. 

Let $Z$ be a random sample of $N$ objects described by features $x_1$ and $x_2$. Table~\ref{tab:prob} gives the probabilities for the four possible combinations of values of the two Bernoulli variables.

%
%

\begin{table}[htb]
	\caption{Notation for the combinations of values of two Bernoulli variables, $w_1$ and $w_2$, in terms of probabilities and counts.}
	\label{tab:prob}
	\centering
	
	\begin{tabular}{cc}
		Probabilities&Counts\\
		\begin{tabular}{rrcc}
			&&\multicolumn{2}{c}{$w_1$}\\
			&&0&1\\
			\cmidrule{3-4}
			\multirow{2}{*}{$w_2$}&0&$a$&$b$\\
			&1&$c$&$d$\\
			\cmidrule{3-4}
			&&&\\
			\multicolumn{4}{c}{$a+b+c+d=1$}
		\end{tabular}
		&
		\begin{tabular}{rrcc}
			&&\multicolumn{2}{c}{$w_1$}\\
			&&0&1\\
			\cmidrule{3-4}
			\multirow{2}{*}{$w_2$}&0&$A$&$B$\\
			&1&$C$&$D$\\
			\cmidrule{3-4}
			&&&\\
			\multicolumn{4}{c}{$A+B+C+D=N$}
		\end{tabular}
	\end{tabular}
\end{table} 

To decide on the sample size $N$ needed to ascertain a difference between $w_1$ and $w_2$, we can apply a McNemar test. The statistic for this test is calculated as 
\begin{equation}
\chi^2 = \frac{(B-C)^2}{B+C}.
\end{equation}
Under the null hypothesis of marginal homogeneity ($p_1=p_2$, also expressed as $b+d=c+d$), $\chi^2$ follows a chi-squared distribution with one degree of freedom. Approximating the counts in Table~\ref{tab:prob} as $N\times$probability, we obtain
\begin{equation}
\chi^2 = \frac{(Nb-Nc)^2}{Nb+Nc}
=N\frac{(p_1-p_2)^2}{p_1+p_2-2d}
\label{chi}
\end{equation}
where $d$ is the probability of both features being correct for a randomly drawn object. Thus $d\leq \min(p_1,p_2)=p_2$ and also $d\geq p_1+p_2-1$. This probability can be thought of as expressing the correlation between the two features. When $x_1$ and $x_2$ are independent, $d=p_1p_2$.

Given $p_1$, $p_2$ and $d$, and choosing the level of significance $\alpha$, we can approximate the necessary sample size $N$ by the inverse of the chi-squared cumulative distribution function $F_\chi(.)$.
We can subsequently rearrange equation~(\ref{chi}) to obtain
\begin{equation}
N = F_\chi^{-1}(1-\alpha)\;\;\frac{p_1+p_2-2d}{(p_1-p_2)^2}\;.
\label{N}
\end{equation}

Clearly, the size will depend on $p_1$ and $p_2$. A large difference $p_1-p_2$ will need fewer objects. The value will also depend on the values of $p_1$ and $p_2$ as well as on the agreement pattern measured by $d$.  

As an example, suppose that we want to estimate the sample size $N$ so that with probability $0.95$ we can discover the difference and pick the correct feature $x_1$. Let $p_1=0.85$ and $p_2=0.80$. Assuming independence, we have $d=p_1p_2=0.68$. Then
\[
N = F_\chi^{-1}(1-0.05)\frac{0.85+0.80-2\times 0.68}{(0.85-0.80)^2}
=445.6\;.
\]

\begin{figure}
	\centering
	\includegraphics[width=0.5\linewidth]{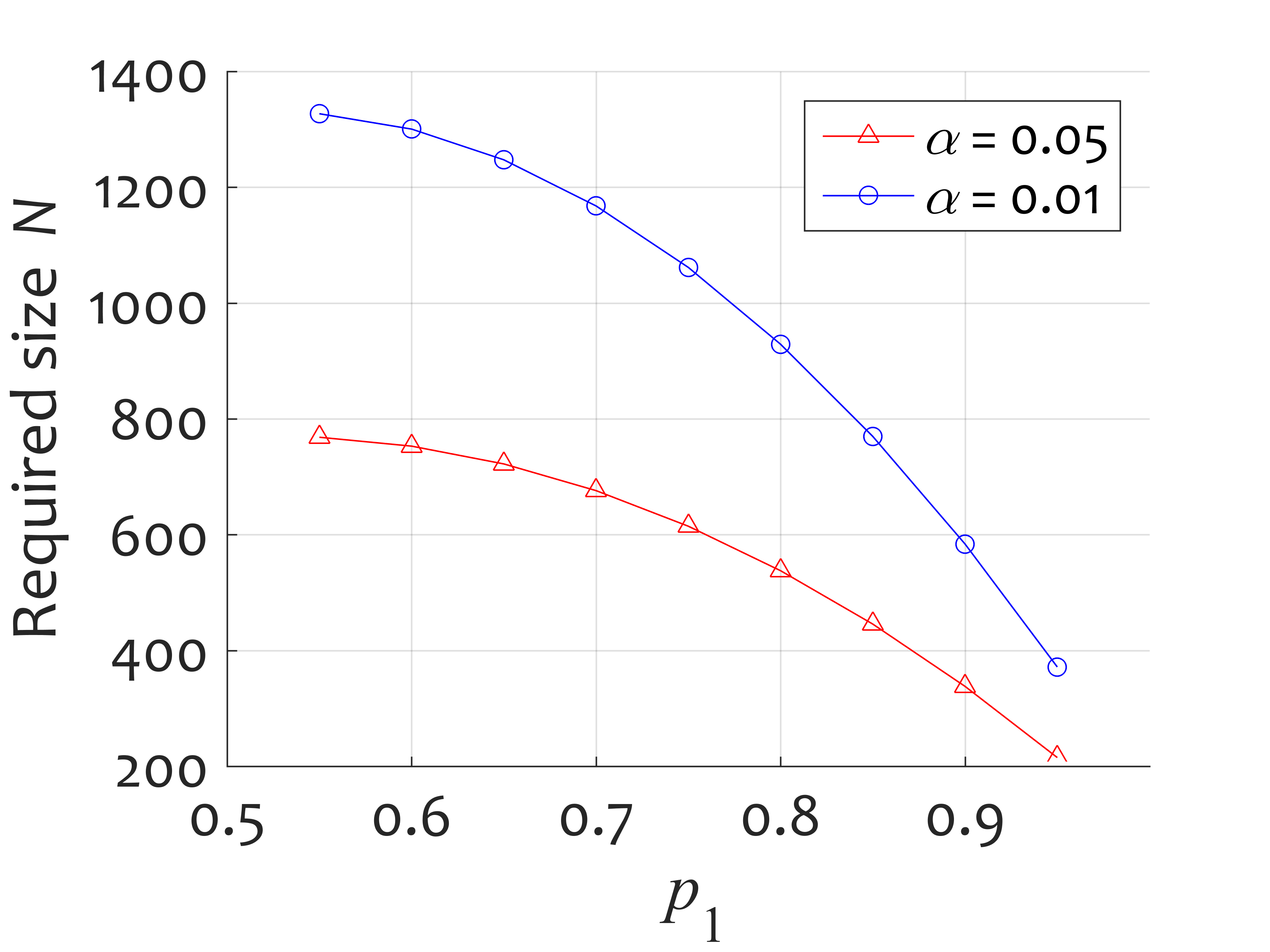}
	\caption{Required size $N$ for picking feature $x_1$ from the pair $(x_1,x_2)$ as a function of $p_1$ for two levels of significance. $p_2 = p_1-0.05$ and $d=p_1p_2$.}
	\label{fig:sizeN}
\end{figure}

Figure~\ref{fig:sizeN} shows the required size $N$ for picking feature $x_1$ from the pair $(x_1,x_2)$. The sizes are plotted as functions of $p_1$ for two levels of significance, $\alpha=0.05$ and $\alpha = 0.01$. For this example we chose $p_2 = p_1-0.05$ and assumed independence between $w_1$ and $w_2$, hence $d=p_1p_2$.

Equation (\ref{N}) shows an interesting effect of the relationship between the Bernoulli variables $w_1$ and $w_2$ on the required sample size. For fixed $\alpha$, $p_1$, and $p_2$, when the agreement between the feature evaluators ($d$) increases, $N$ decreases. In other words, for larger agreement, fewer observations are needed to pick feature $x_1$ with the same certainty. Indeed, if we take the maximum possible agreement $d=p_2=0.80$ in the above example, this leaves $c=0$ and $b=0.05$. In this case, Eqn.(\ref{N}) returns $N = 77$. While some dependency between $w_1$ and $w_2$ is expected, the sample size required to pick confidently {\em between just two features} is very large. Exactly the same argument is valid when we compare two {\em sets} of features. However, generalising this result to sets of more than two elements (thousands of features!) will amount to quite large values of $N$. Even mitigated by high dependency between the variables $w_i$, these sample counts expose the inadequacy of feature selection from very small samples, which reinforces the doubts and warnings published elsewhere~\cite{EinDor06,Varoquax18}.

\subsection{Smoothed estimator}

Assume now that instead of the binary variable $w_i$ for feature $x_i$, we have a continuous-valued variable $v_i$ estimating the probability of correct classification for the value of $x_i$ of the given object, where the classifier is built on feature $X_i\in X$ alone. Denote this probability by $P(C_i|x_i)$. Unfortunately, we cannot assume that such probabilities would adhere to a normal distribution. Assume again that $x_1$ is better than $x_2$ ($p_1=E[v_1]>p_2=E[v_2]$). In this case, a Wilcoxon signed-rank test can be used. This non-parametric test will have less statistical power than a parametric test such as the t-test. Thus, the Wilcoxon test may not be able to detect an existing difference for normally distributed variables, which would be detectable by the parametric test. This implies that larger samples would be needed in the same set-up to detect the difference through the non-parametric test. Then, we can use the normal distribution assumption to illustrate the magnitude of $N$, noting that this is a lower estimate.

%
%

Assuming $v_1$ and $v_2$ both have a normal distribution,
a paired t-test can be applied to ascertain that $p_1>p_2$. For large $N$ (which we expect here), the t-distribution approaches the normal distribution. Denote by $v$ the difference $v=v_1-v_2$.  Assuming that $v_1$ and $v_2$ are jointly normally distributed, variable $v$ has normal distribution too with mean $\mu_v=p_1-p_2$ and variance $\sigma_v=\sigma_1^2+\sigma_2^2-2\sigma_{12}$, where $\sigma_i^2$ is the variance of $v_i$, $i=1,2$, and $\sigma_{12}$ is the covariance between $v_1$ and $v_2$. If we measure $\bar{v}$ as an average from sample $Z$ of size $N$, we can use the cumulative distribution function of the standard normal distribution $\Phi$ to estimate critical values 
for a chosen level of significance $\alpha$, and estimate the required values of $N_c$ for the continuous-valued case\footnote{For the normal distribution we can use only one tail, which implies that we can replace $\alpha$ in Eqn.(\ref{Nnormal}) with $\alpha/2$, leading to an even higher requirement for $N$. }
\begin{equation}
N_c = \Phi^{-1}(1-\alpha)\;\;\frac{\sigma_1^2+
	\sigma_2^2-2\sigma_{12}}{(p_1-p_2)^2}\;.
\label{Nnormal}
\end{equation}

This equation is largely similar to Eqn.(\ref{N}), and leads to the same conclusions about the considerable required sample sizes and the mitigating effect of positive covariance between the variables estimating the features' merit.

%

If ascertaining difference between two features requires sample sizes of this magnitude, selecting more features would require many thousands of instances. This resonates with the previous studies, and gives an early argument for our cautionary tale.

\section{Low correlation between estimated and true error}
\label{rel}

To define the problem formally, consider a labelled dataset $Z=\{(\z_1,y_1),$ $\ldots, (\z_N,y_N)\}$ where $\z_j\in \bb{R}^n$ are objects represented as points in some $n$-dimensional space, and $y_j\in \Omega=\{\omega_1,\ldots,\omega_c\}$ are class labels, $j=1,\ldots,N.$ The dataset is drawn randomly from a joint distribution $p(\x,y)$. The task of feature selection is to reduce the feature space $\bb{R}^n$ to a space of a much lower dimensionality $d \ll n$ by dropping dimensions of the original space.   

\subsection{Types of error in feature selection from a finite sample}
Let $X=\{X_1,X_2,...,X_n\}$ be the feature set. The first issue we consider is what we measure and return as the result of the feature selection. Assume that a classifier model $C$ has been chosen for this problem. Denote by $y$ the true label of instance $\x$. Let $C(\x|Z)$ be the output of the classifier trained on dataset $Z$ for instance $\x$. To compact the notation, let $Z_a^b$ be a dataset $Z$ of cardinality $a$ and containing only feature set $b\subseteq X$. Denote by $\bb{E}$$[\zeta]$ an operator which calculates the expectation of $\zeta$ over the respective feature space and probability distribution.
Assume also that we have access to the whole population of interest, denoted $Z_{\infty}^X$. For a selected $S\subseteq X$, consider the probability that $C$ is correct, denoted $Pr\left(C(\x|Z_{\infty}^S)\neq y\right)$. Classifier $C$ is fixed, which means that it will output only one class label for $\x$. The uncertainty, measured by this probability, comes from the fact that $\bf x$ may originate from different classes with probability $P(y = \omega_i|\x)$, $i=1,\ldots,c$. Then the error that we would like to return to the user together with feature subset set $S$, {\em given the chosen classifier $C$}, is

	\begin{equation}
	e(S,C) = \bb{E}\left[Pr\left(C(\x|Z_{\infty}^S)\neq y
	\right)\right]
	= \int_{\x'\in\bb{R}^{|S|}} Pr\left(C(\x|Z_{\infty}^S)\neq y\right) \; p(\x') \; d\x'\;\;\;,
	\end{equation}

$y$ is the true label of $\x$, $\x'$ is the restriction of $\x$ using only the features in set $S$, and $p(\x')$ is its probability density function marginalised across $X\setminus S$. Ideally, we would like to return $e$ to the user together with $S$ but in reality we have only some estimate $\hat e$ (Table~\ref{tab:errortype}). 

Feature selection algorithms look for the {\em optimal} feature subset $S^*\subseteq X$ for the problem at hand and for the chosen $C$
\begin{equation}
S^*= \arg\min_S \left\{e(S,C)\right\}\;,
\end{equation}
and hence $e^*= e(S^*,C)$ is the true error of the best subset, given classifier $C$ (Table\ref{tab:errortype}). We can remark here that finding this error is predicated upon having access to the whole population of interest, as well as being able to identify $S^*$. The latter assumes that we have a perfect feature selection algorithm at hand. 

Suppose that we can only afford to sample $Z$ of cardinality $N$, i.e., $Z_N^X$. Theoretically, a classifier trained on a finite sample will be less accurate than the one trained on $Z_{\infty}^X$. Furthermore, the optimal subset of features $S^*$ may not be identifiable from the sampled $Z_N^X$. Instead, we may have another optimal set $S_N^*\subseteq X$. It is possible that $S^*\neq S_N^*$ because the classifier trained on the smaller sample may not be identical to the ideal classifier, and will be either equivalent or more likely inferior to it. In the absence of an infinite dataset, the purpose of feature selection is to return to the user $S_N^*$ and the respective error $e_N^*$ (Table\ref{tab:errortype}).

Furthermore, our feature selection algorithm may not be able to identify the best subset for the sampled dataset. The reason is that the {\em estimates} of the errors from such a small dataset typically have a very large variance. Therefore, we cannot measure the true error of each feature subset, and will likely return  a non-optimal subset $S$ and its error $\hat e$. The inferior subset may come as a result of either a suboptimal feature selection procedure (e.g., a sequential forward selection) of a poor estimate of the accuracy of the subset, so that even exhaustive search will return a suboptimal $S$. Most likely, the inferior result is due to the combination of both.

\begin{table}[htb]
	\caption{Type of errors and related feature sets and data sources.}
	\label{tab:errortype}
	\[
	\begin{array}{@{}rccc@{}}
	\midrule
	\mbox{Description}&\mbox{Notation}&\mbox{Feature set}&\mbox{Data source}\\
	\midrule
	\mbox{Best feature subset}&e^*&S^*&Z_{\infty}^X\\
	\mbox{Sample best feature subset}&e_N^*&S_N^*&Z_N^X\\
	\mbox{Returned to user}&\hat e&S~&Z_N^X\\
	\midrule
	\end{array}
	\]
\end{table}

By the argument of optimality, we have:
\begin{equation}
e^*\leq e_N^* \leq e.
\end{equation}

However, $\hat e$ could be anywhere with respect to the above errors. In a perfect scenario, $S^*=S$ and $e_N^*= e = \hat e$. Our main argument in this study is that, for wide datasets, $\hat e$ is a dangerously poor estimate of $e$, leading to spurious choices of $S$ with misleading estimates $\hat e$.

\subsection{Error estimators (finite testing sample)}

To select $S$ and calculate the estimate of $e$ we re-test the classifier on $Z$ using some acceptable training-testing protocol. Thus, instead of the true $e$, the returned quantity will be an estimate, sometimes quite far off.
The resubstitution estimate $\hat e^R$ evaluates the error on the training data and is known to be optimistically biased~\cite{Devroye96,Kim09}. The leave-one-out error $\hat e^{LOO}$ applied on $Z$ {\em after} selecting $\hat S$ will likely be also optimistically biased, more so for smaller datasets~\cite{KunchevaPR18}. The `right' protocol requires a subset of features is selected and evaluated from each cross-validation fold. The error is estimated as the average of the errors in the individual folds. A 10-fold LOO will result in 10 different subsets $S$. Therefore, after $\hat e$ is estimated, a single $S$ is selected using the whole of $Z$. This protocol gives an estimate $\hat e^{r-LOO}$ which was found to be only slightly pessimistically biased~\cite{KunchevaPR18}. 
{\em On average} across different samples $Z$, we expect
\begin{equation}\\
{\hat e}^R\leq {\hat e}^{LOO}\leq {\hat e}^{r-LOO}.
\end{equation}

It has been argued that the quality of the estimate of the error criterion is  more important in feature selection than the feature selection method itself~\cite{Sima05b}.  Various estimators have been proposed over the years to complement the conventional resubstitution and cross-validation estimators. Among the alternative estimators are the bootstrap estimator~\cite{Efron97}, the bolstered estimator~\cite{BragaNeto04,Sima05a}, and smoothed estimators~\cite{Glick78}, the latter being specifically useful for small datasets~\cite{Knocke86}. Comparisons between the estimators~\cite{Dougherty10,Kim09} reveal that the time-consuming ones such as bootstrap 632+ and the bolstered estimator have an edge over the competitors but the general relationship between the true error and the estimated error is poor~\cite{Dougherty10,BragaNeto10}. While most of these findings are based on simulated data, there is no reason to expect much different results with real data. 

In this study we chose the smoothed error estimator because it is computationally more efficient than bootstrap and bolster estimators, and dramatically reduces the number of ties in comparison with the counting estimator. The smoothed estimator relies on the ability of the classifier model to output estimates of posterior probabilities for the classes. In other words, instead of recording 0 (correct classification) and 1 (error), for each $\x$ we store the estimate of the probability of predicting an incorrect class. These probabilities are subsequently averaged over the testing instances to arrive at a single estimate of the error. Arguably, the quality of the estimates of the probabilities will suffer from the low sample size as well. Nonetheless, the smoothed estimator gives us better flexibility in choosing the feature subset. 

\subsection{An example}
Here we illustrate the difficulty of feature selection from wide datasets using an example from the work by Reunanen, 2003~\cite{Reunanen03}. A small sample was taken from the `sonar' dataset from the UCI repository~\cite{UCI} as $Z$ with 10 instances from each of the two classes. The rest of the dataset (188 instances) was used as the unseen testing data, from which we estimate a version of the `true' error. The data has two balanced classes (47\%/54\%) and 60 numerical features.

First we rank the features using the Symmetric Uncertainty ranker~\cite{Hall03}. Next we evaluate all 1\,023 combinations of the top 10 features ($2^{10}$, excluding the empty set) with respect to the linear discriminant classifier (LDC). Four estimates were calculated: (i) ${\hat e}^R$ Resubstitution error (LDC trained and tested on $Z$); (ii) ${\hat e}^{LOO}$ Leave-One-Out (LOO) error on $Z$; (iii) ${\hat e}^{s-LOO}$ Smoothed LOO error on $Z$; and (iv) ${\hat e}'$ `True' error (LDC trained on $Z$ and evaluated on the testing set). 

\begin{figure*}
	\centering
	\includegraphics[width=\linewidth]{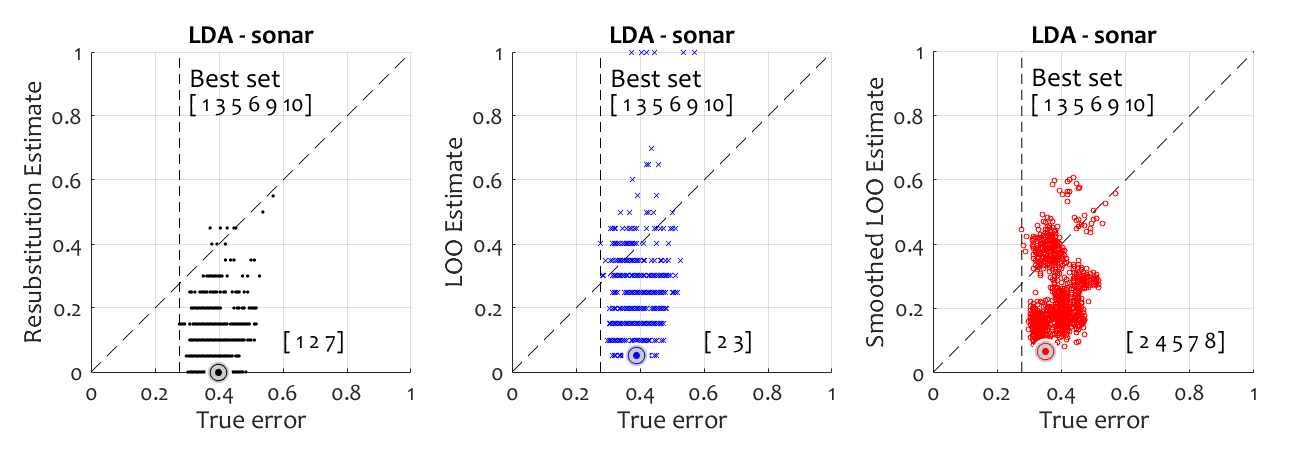}
	\caption{Scatterplot of the error estimates ${\hat e}^R$, ${\hat e}^{LOO}$, and ${\hat e}^{s-LOO}$ versus the `true' error ${\hat e}'$ for the sonar dataset. Each plot contains 1\,023 points, one for each subset of the top 10 features. The error estimates are calculated from a single sample $Z$ of 20 instances (10 per class).}
	\label{illu}
\end{figure*}

Figure~\ref{illu} shows the first set of results. In each plot, an estimator (i)-(iii) is plotted against the true error (iv). Each feature subset is a point on the plot, hence there are 1\,023 points in each plot. If the estimator from $Z$ was ideal, all points would lie on the diagonal shown in each plot. 

The array next to each cloud of points contains the indices of the features in the best set according to the respective error estimator. The features are numbered from 1 to 10 according to their rank in the list. If there was a tie at the minimum error (nearly inevitable for ${\hat e}^R$ and ${\hat e}^{LOO}$), the sets of minimum cardinality were identified among the tied sets, and a random set was chosen among these. The best set according to ${\hat e}'$ is shown in each plot for comparison.

This illustration eliminates the effect of a possible sub-optimal search because we have carried out an exhaustive search on the top 10 features. The dramatic discrepancy between the feature sets found by the estimators and the true best set are solely owed to the inadequate error estimates. As seen in the plot (reinforcing finding in past literature~\cite{BragaNeto10}), there is no visible relationship between the true error and its estimates. Thus, not only is the returned feature subset a poor match of the best set, but the error estimate predicted by the estimator is highly optimistically biased for this $Z$. Table~\ref{tab:illu} shows the error rates for the three estimators.

\begin{table}
	\caption{Error rates for the selected feature sets}
	\label{tab:illu}
	\centering
	\begin{tabular}{@{}rllcc@{}}
		\midrule
		Estimator&&Feature set&Predicted&True\\
		\midrule
		Resubstitution&${\hat e}^R$&[1,2,7]&0.0000&0.3989\\
		LOO& ${\hat e}^{LOO}$&[2,3]&0.0500&0.3883\\
		Smoothed LOO&${\hat e}^{s-LOO}$&[2,4,5,7,8]&0.0662&0.3511\\
		\midrule
		True&${\hat e}'$&[1,3,5,6,9,10]&--&0.2766\\
		\midrule
		All features&--&[1--10]&--&0.3830\\
		\hline
	\end{tabular}
\end{table}

We should note that the discrepancy between the returned feature sets identified by different estimators may not be entirely the estimators' fault. It is possible that low error rate is achievable by more than one feature subset. The value of the estimate, however, should be a good approximation of the true error, which was not observed in the experiment. 

This example suggests a slight advantage of the smoothed LOO estimator over the standard LOO. First, the chances of a tie are greatly reduced, and, second, only this estimator led to a smaller true error than that using all 10 top features. Still, the main message here is the evident lack of relationship between predicted and true error for very low-sample-size datasets.

\subsection{Search method (suboptimal search)}
Suppose that we have an oracle error estimator which correctly predicts the generalisation error rate of a classifier trained on the given dataset $Z$. It is well known that the exhaustive search through all subsets of features is the only method guaranteeing that the set with the minimum error will be found~\cite{Cover77}. Due to its computational intensity ($2^n-1$ repeats of training a classifier and evaluating its error) this approach is feasible only for small cardinality of the feature space, for example, up to 15 features. A simpler alternative approach is a ``heavy'' random search whereby random feature subsets are sampled and evaluated. More intricate procedures such as floating forward selection (SFFS)~\cite{Pudil94} and evolutionary algorithms~\cite{Xue16} are also possible if the initial feature set is reduced to a smaller set as a preliminary step.

In summary, imperfect error estimation in combination with a non-optimal search procedure may result in selecting a spurious feature set with an inadequate error estimate. This problem is exacerbated by the small sample size. In this paper we argue that, for wide datasets, we should stop at ranking the features in $X$ and returning the subset of the top $d$ features. Search procedures to reduce this set further (a wrapper approach) most often lead to over-using the data and reaching arbitrary conclusions. 

\subsection{Classifier dependence}
Note that in the preceding discussion, the classifier model $C$ is chosen and fixed. This brings us to the classifier issue: feature selection is intrinsically classifier-dependent. A classifier-independent approach~\cite{Abe06} may work when the classes are clearly separable in some feature subspace so that any classifier model will find the separation ``easy''. However, this approach may overlook highly useful feature subspaces which favour a specific classifier model. 

\begin{figure}
	\centering
	\begin{tabular}{cc}
		\includegraphics[width=0.45\linewidth]{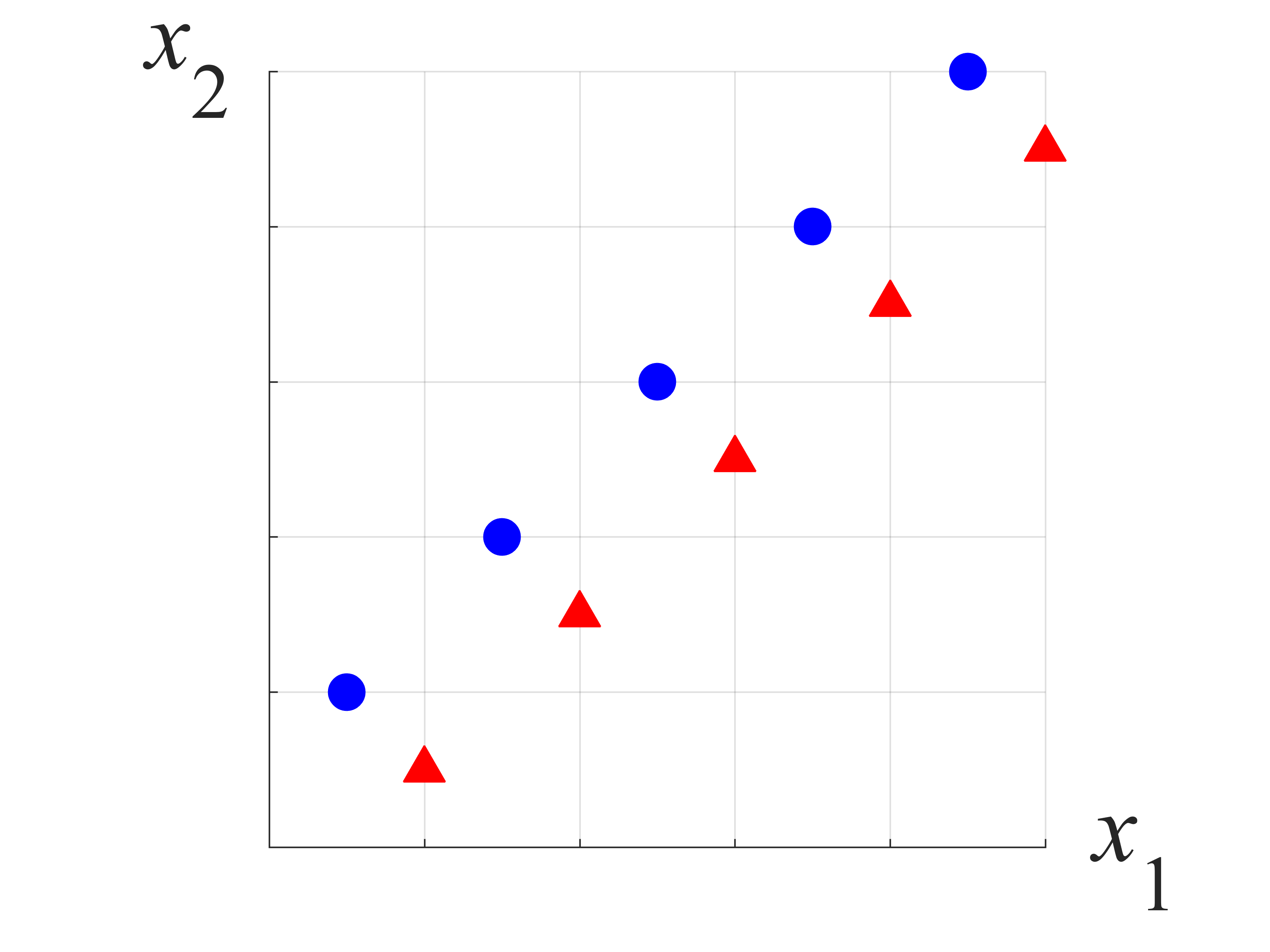}&
		\includegraphics[width=0.45\linewidth]{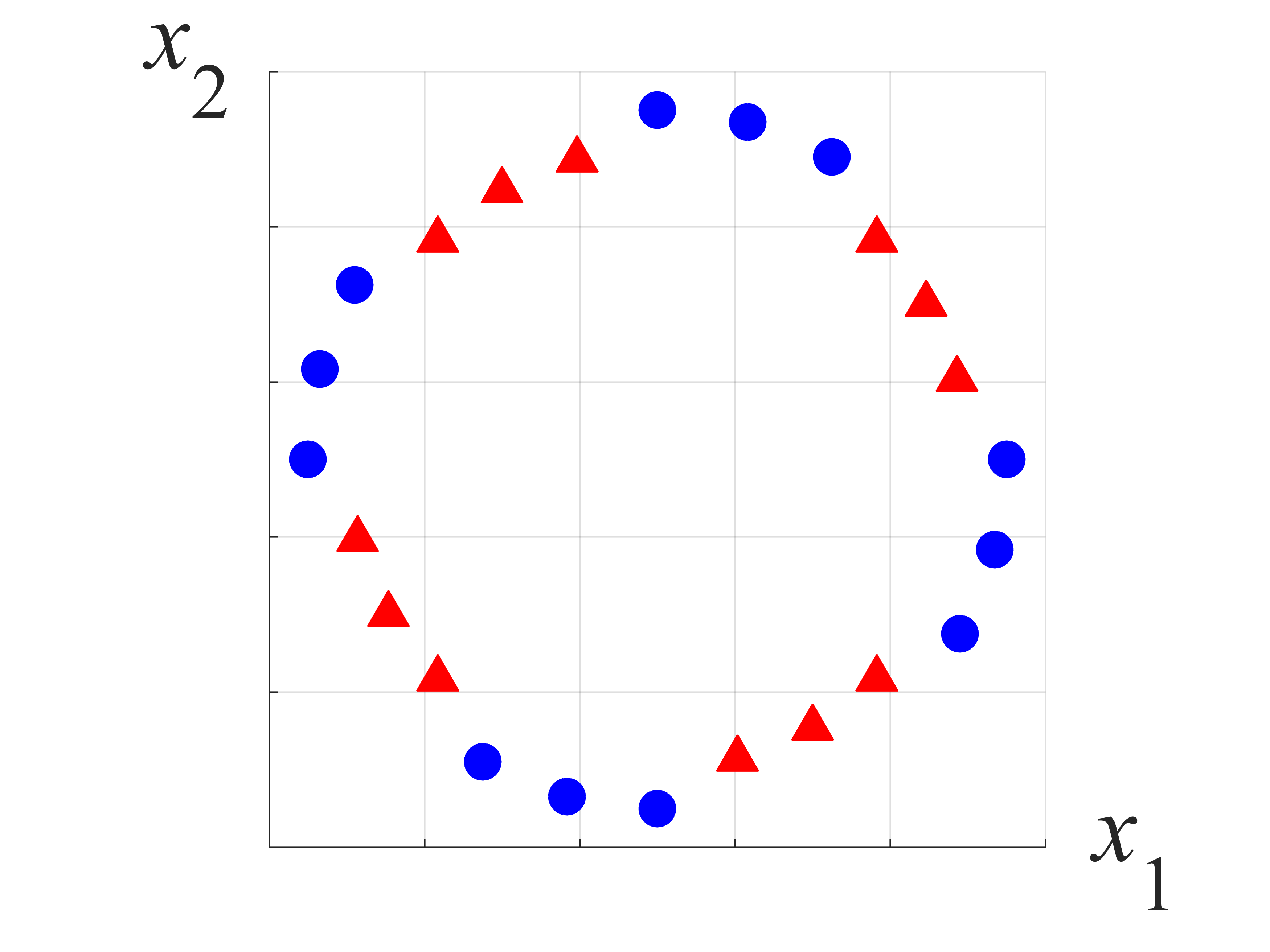}\\
		(a) $\hat e^{LOO}(\mbox{\scriptsize LDC}) = 0$&(b) $\hat e^{LOO}(\mbox{\scriptsize LDC}) = 1$\\
		$\hat e^{LOO}(\mbox{\scriptsize 1NN}) = 1$&$\hat e^{LOO}(\mbox{\scriptsize 1NN}) = 0$\\
	\end{tabular}
	\caption{Examples of points in two-dimensional feature spaces coming from two classes depicted with different markers. While the features are useless individually, each pair is suitable for a different classifier model: (a) gives 0\% LOO error for LDC, and (b), for 1NN.}
	\label{fig:classifier-dependent}
\end{figure}

An example is shown in Figure~\ref{fig:classifier-dependent}.  The two features, assessed individually, are nearly useless. However, the features in plot (a) give together zero LOO error for the linear discriminant classifier (LDC), and LOO error 1 (all objects are classified incorrectly) for the nearest neighbour classifier (1NN). The opposite case is shown in plot (b).\footnote{The MATLAB code for generating the example datasets and calculating the LOO errors is available at \url{https://github.com/LucyKuncheva/Feature-Selection}. } The choice of a classifier model will ultimately determine the importance of the pair of features.

In this study we will use seven classifiers that broadly cover the spectrum of most used classifiers in wide datasets.
Even though different classifiers have different sensitivity to changing the feature space, we will assume that, in general, our conclusions will generalise across different classifier models.

\section{Experiments}
\label{exp}
In this experiment, we artificially simulate very low-sample data sets. To this end, we sample only {10 objects per class} as our data set $Z_{10}^X$. This is an extreme scenario which allows us to demonstrate the issues of concern listed in the introduction. A perfect collection of datasets would contain only {\em large} data sets, that is, data of high dimensionality and a large sample size. Since our sample size is very small, we would have large remaining data to use in place of $Z_{\infty}^S$. Unfortunately, we could not identify a sufficiently large collection of large datasets, which led us to using some wide data sets too. Nonetheless, since our $N=10c$ is extremely small, a reasonably-sized data for testing is left even in the relatively wide datasets in our experiment.

\subsection{Datasets}
Previous studies were mostly confined to simulated data with known (and well-behaved) distributions. Here we use 19 real and 1 artificial\footnote{Madelon is an artificial dataset, nonetheless it is a complex problem because is multivariate and highly non-linear.} datasets sourced from \url{http://featureselection.asu.edu/datasets.php}~\cite{Li16}. All datasets were chosen so that we have a sufficient number of samples to allow for calculating a reasonable estimate of the true error $e$ after cutting out a sample $Z$ of size $N$. Where the dataset contained more than two classes, we selected the two most frequent classes. Table~\ref{tab:data} shows the properties of the datasets. 

\begin{table}[htb]
	\caption{Characteristics of the high-dimensional datasets.}
	\label{tab:data}
	\centering
	
	\smallskip
	\begin{tabular}{@{}rlrr@{}}
		\toprule
		Number & Dataset &  Instances &  Features \\
		\midrule
		1 & ALLAML        & 72     & 7\,129 \\   
		2 & arcene        & 200    & 10\,000 \\  
		3 & BASEHOCK      & 1\,993 & 4\,862 \\   
		4 & Carcinom      & 53     & 9\,182 \\   
		5 & CLL\_SUB\_111 & 100    & 11\,340 \\  
		6 & COIL20        & 144    & 1\,024 \\   
		7 & colon         & 62     & 2\,000 \\   
		8 & gisette       & 7\,000 & 5\,000 \\   
		9 & GLI\_85       & 85     & 22\,283 \\  
		10& Isolet        & 120    & 617 \\      
		11& leukemia      & 72     & 7\,070 \\   
		12& lung          & 160    & 3\,312 \\   
		13& madelon       & 2\,600 & 500 \\      
		14& PCMAC         & 1\,943 & 3\,289 \\   
		15& Prostate\_GE  & 102    & 5\,966 \\   
		16& RELATHE       & 1\,427 & 4\,322 \\   
		17& SMK\_CAN\_187 & 187    & 19\,993 \\  
		18& TOX\_171      & 90     & 5\,748 \\   
		19& USPS          & 2\,822 & 256 \\      
		20& warpPIE10P    & 42     & 2\,420 \\   
		\bottomrule
	\end{tabular}
	
\end{table}

\subsection{Methods}
For each dataset we repeated the following procedure 10 times (``sampling runs''): 
\begin{enumerate}
	\item Choose a classifier $C$ and a ranker method $R$.
	\item Take a random sample $Z$ from the dataset containing 10 instances from each class.
	\item Rank the features in $Z$ by the ranker method $R$, and store the top 20 features in set $S$.
	\item Apply each of the Selection Schemes from the list below. 
\end{enumerate}

Using the chosen classifier $C$ and the smoothed leave-one-out estimator as the evaluation criterion ($e^{s-LOO}$), we applied the Selection Schemes listed below. Shown in parentheses is the number of required evaluations of the $LOO$ error. Note that each such evaluation requires $N$ training and testing iterations.

\medskip
\noindent
List of selection schemes:

\smallskip
\begin{enumerate}[label=\alph*.]
	\item {\em ALL.} All features in the original dataset. (1) 
	\item {\em Top3.} Top three features from the ranked list. (1)
	\item {\em Top10.} Top 10 features. (1)
	\item {\em Top20.} Top 20 features. (1)
	\item {\em Best3.} Best three features selected through enumerating all combinations of three features out of the top 20. $\left(\left({20\atop 3}\right) = 1140\right)$
	\item {\em EX10.} Best subset found through exhaustive search among all subsets of the top 10 features. $\left(2^{10} = 1024\right)$ Note: For the empty set we used the prior probabilities estimated from the sample.
	\item {\em RND20.} Best subset identified through random search from the 20 features. 1024 evaluations  were carried out to tally with EX10. At each evaluation, a feature subset was created switching each feature from the top 20 `on' or `of' with probability 0.5. (1024) 
\end{enumerate}

\medskip
The classifier models $C$ which we used here were:

\begin{enumerate}
	\item Nearest neighbour (1NN).
	\item Decision tree (DT).
	\item Linear discriminant classifier (LDC).
	\item Na\"ive Bayes (NB).
	\item Random Forest (RF).
	\item Support Vector Machine with a Gaussian kernel (SVMG).
	\item Support Vector Machine with a linear kernel (SVML).
\end{enumerate}

We ran the experiments in the WEKA environment~\cite{Hall09}.
To calculate the smoothed estimator we used the posterior probabilities as provided by WEKA. The exception is 1NN where we applied the softmax transformation, it gives smoother probabilities.
In WEKA, for DT the probabilities are obtained using the class distribution of the training examples in the corresponding leaf. For LDC, the probabilities are obtained using  Bayes' rule on the estimated Gaussians. 
For NB, the probabilities for the attributes are obtained from the normal distribution. For RF, the probabilities are the averages of the trees probabilities. For SVMG and SVML with two classes, the probabilities are discrete, in $\{0, 1\}$.


The WEKA implementation of five ranking methods were used for determining the top 20 features: 

\begin{enumerate}[label=\Alph*.]
	\item Symmetrical Uncertainty~\cite{Hall03}.
	\item Random Forest~\cite{Breiman01}.
	\item ReliefF~\cite{Kira92,Robnik03}.
	\item SVM~\cite{Guyon02}.
	\item SVMRFE~\cite{Guyon02}.
\end{enumerate}

\medskip
Thus, for each dataset we ran [7 (classifiers) $\times$ 5 (rankers) $\times$ 6 (selectors) $\times$ 10 (sampling runs)] $+$ [7 (classifiers) $\times$ 1 (selector = ALL) $\times$ 10 (sampling runs)] = 2\,170  experiments. A full set of results is presented in the supplementary material.

\subsection{Results}

The main result from our experiment is presented in Table~\ref{tab:res}. For a given combination of classifier and ranker, $<C,R>$, we collated the errors  $e^{s-LOO}$ for the 10 sampling runs for each dataset produced by the 7 selectors. Thus, we constructed a matrix of size 20 (datasets) $\times$ 10 (sampling runs) = 200 rows, and 7 (selectors) columns. We are interested in comparing the selectors across datasets and sampling runs. Since the errors may not be commensurable across different datasets, we calculated the ranks of the selectors for each row of the table. The best selector method received rank 1, and the worst, rank 7. The averaged ranks across datasets and runs are shown in the table. We also colour-coded the cells of the table. Bright red corresponds to the best selectors and dark blue, to the worst. The best result in each row is highlighted in boldface.


For each $<C,R>$ combination we ran the Friedman test (a non-parametric version of balanced two-way ANOVA) to determine whether the differences between the selectors were significant. As we have 10 runs for the same dataset, the block version of the test was applied\footnote{We used the MATLAB Statistics Toolbox implementation of the Friedman test.}.  

\begin{table}[p]
	\caption{Average ranks for the 7 Selectors for all combinations of Classifier and Ranker. The best Selectors are shaded in bright red, and the worst, in dark blue. The best selector in each row is shown in a box, and in boldface. The selectors which are indistinguishable from the best one of the row, according to the Friedman test, are given in boxes too.}
	\label{tab:res}
	\centering
	\renewcommand{\arraystretch}{1.2}
	{\footnotesize
		\begin{tabular}{llccccccc}
			&&\multicolumn{7}{c}{\em SELECTOR}\\
			\cmidrule{3-9}
			\multicolumn{1}{c}{\em C} &\multicolumn{1}{c}{\em R}& Best3&
			EX10&
			RND20&
			Top3&
			Top10&
			Top20&
			ALL\\
			\cmidrule{1-9}
			\multirow{5}{*}{1NN}
			&RF& \cellcolor[rgb]{0.91,0.91,1.00} 4.12 &\cellcolor[rgb]{0.84,0.84,1.00} 4.26 &\cellcolor[rgb]{1.00,0.61,0.61} 3.47 &\cellcolor[rgb]{0.00,0.00,1.00} 5.54 &\cellcolor[rgb]{0.95,0.95,1.00} 4.08 &\cellcolor[rgb]{1.00,0.57,0.57} 3.42 &\cellcolor[rgb]{1.00,0.37,0.37} \fbox{\bf 3.12} \\
			&RelF& \cellcolor[rgb]{0.67,0.67,1.00} 4.59 &\cellcolor[rgb]{0.95,0.95,1.00} 4.08 &\cellcolor[rgb]{1.00,0.57,0.57} \fbox{3.42} &\cellcolor[rgb]{0.78,0.78,1.00} 4.39 &\cellcolor[rgb]{1.00,0.79,0.79} 3.77 &\cellcolor[rgb]{1.00,0.50,0.50} \fbox{\bf 3.30} &\cellcolor[rgb]{0.76,0.76,1.00} 4.45 \\
			&SVM& \cellcolor[rgb]{0.70,0.70,1.00} 4.53 &\cellcolor[rgb]{1.00,0.94,0.94} 3.93 &\cellcolor[rgb]{1.00,0.62,0.62} \fbox{3.48} &\cellcolor[rgb]{0.45,0.45,1.00} 4.79 &\cellcolor[rgb]{0.96,0.96,1.00} 4.04 &\cellcolor[rgb]{1.00,0.44,0.44} \fbox{\bf 3.23} &\cellcolor[rgb]{1.00,1.00,1.00} 4.00 \\
			&RFE& \cellcolor[rgb]{0.74,0.74,1.00} 4.49 &\cellcolor[rgb]{1.00,0.90,0.90} 3.88 &\cellcolor[rgb]{1.00,0.56,0.56} \fbox{3.40} &\cellcolor[rgb]{0.44,0.44,1.00} 4.80 &\cellcolor[rgb]{1.00,0.82,0.82} 3.82 &\cellcolor[rgb]{1.00,0.41,0.41} \fbox{\bf 3.19} &\cellcolor[rgb]{0.77,0.77,1.00} 4.42 \\
			&SU& \cellcolor[rgb]{0.62,0.62,1.00} 4.64 &\cellcolor[rgb]{1.00,0.81,0.81} 3.82 &\cellcolor[rgb]{1.00,0.67,0.67} \fbox{3.58} &\cellcolor[rgb]{0.57,0.57,1.00} 4.68 &\cellcolor[rgb]{1.00,0.73,0.73} 3.71 &\cellcolor[rgb]{1.00,0.48,0.48} \fbox{\bf 3.29} &\cellcolor[rgb]{0.83,0.83,1.00} 4.29 \\
			\cmidrule{1-9}
			\multirow{5}{*}{DT}
			&RF& \cellcolor[rgb]{1.00,0.98,0.98} \fbox{3.97} &\cellcolor[rgb]{0.94,0.94,1.00} \fbox{4.08} &\cellcolor[rgb]{1.00,0.79,0.79} \fbox{3.77} &\cellcolor[rgb]{0.48,0.48,1.00} 4.76 &\cellcolor[rgb]{1.00,0.92,0.92} \fbox{3.91} &\cellcolor[rgb]{1.00,0.93,0.93} \fbox{3.92} &\cellcolor[rgb]{1.00,0.68,0.68} \fbox{\bf 3.59} \\
			&RelF& \cellcolor[rgb]{0.97,0.97,1.00} \fbox{4.04} &\cellcolor[rgb]{1.00,0.85,0.85} \fbox{\bf 3.85} &\cellcolor[rgb]{0.99,0.99,1.00} \fbox{4.01} &\cellcolor[rgb]{1.00,0.86,0.86} \fbox{3.86} &\cellcolor[rgb]{1.00,0.95,0.95} \fbox{3.93} &\cellcolor[rgb]{1.00,0.91,0.91} \fbox{3.89} &\cellcolor[rgb]{0.77,0.77,1.00} 4.42 \\
			&SVM& \cellcolor[rgb]{0.86,0.86,1.00} 4.20 &\cellcolor[rgb]{0.90,0.90,1.00} \fbox{4.16} &\cellcolor[rgb]{1.00,0.92,0.92} \fbox{3.92} &\cellcolor[rgb]{0.92,0.92,1.00} \fbox{4.11} &\cellcolor[rgb]{1.00,0.95,0.95} \fbox{3.94} &\cellcolor[rgb]{1.00,0.70,0.70} \fbox{\bf 3.65} &\cellcolor[rgb]{0.98,0.98,1.00} \fbox{4.04} \\
			&RFE& \cellcolor[rgb]{0.91,0.91,1.00} \fbox{4.13} &\cellcolor[rgb]{1.00,0.91,0.91} \fbox{3.90} &\cellcolor[rgb]{1.00,0.99,0.99} \fbox{3.98} &\cellcolor[rgb]{1.00,0.97,0.97} \fbox{3.96} &\cellcolor[rgb]{1.00,0.87,0.87} \fbox{3.86} &\cellcolor[rgb]{1.00,0.86,0.86} \fbox{\bf 3.85} &\cellcolor[rgb]{0.82,0.82,1.00} \fbox{4.31} \\
			&SU& \cellcolor[rgb]{0.92,0.92,1.00} \fbox{4.11} &\cellcolor[rgb]{1.00,0.98,0.98} \fbox{3.97} &\cellcolor[rgb]{0.98,0.98,1.00} \fbox{4.04} &\cellcolor[rgb]{1.00,0.76,0.76} \fbox{\bf 3.73} &\cellcolor[rgb]{1.00,0.88,0.88} \fbox{3.86} &\cellcolor[rgb]{1.00,0.89,0.89} \fbox{3.87} &\cellcolor[rgb]{0.76,0.76,1.00} 4.42 \\
			\cmidrule{1-9}
			\multirow{5}{*}{LDC}
			&RF& \cellcolor[rgb]{1.00,0.78,0.78} 3.76 &\cellcolor[rgb]{1.00,1.00,1.00} 4.00 &\cellcolor[rgb]{0.71,0.71,1.00} 4.52 &\cellcolor[rgb]{0.50,0.50,1.00} 4.75 &\cellcolor[rgb]{0.86,0.86,1.00} 4.20 &\cellcolor[rgb]{0.13,0.13,1.00} 5.36 &\cellcolor[rgb]{1.00,0.00,0.00} \fbox{\bf 1.42} \\
			&RelF& \cellcolor[rgb]{1.00,1.00,1.00} 3.99 &\cellcolor[rgb]{1.00,0.89,0.89} 3.87 &\cellcolor[rgb]{0.31,0.31,1.00} 4.96 &\cellcolor[rgb]{1.00,0.80,0.80} 3.78 &\cellcolor[rgb]{0.96,0.96,1.00} 4.06 &\cellcolor[rgb]{0.09,0.09,1.00} 5.52 &\cellcolor[rgb]{1.00,0.18,0.18} \fbox{\bf 1.81} \\
			&SVM& \cellcolor[rgb]{0.83,0.83,1.00} 4.28 &\cellcolor[rgb]{1.00,0.83,0.83} 3.83 &\cellcolor[rgb]{0.40,0.40,1.00} 4.83 &\cellcolor[rgb]{0.89,0.89,1.00} 4.17 &\cellcolor[rgb]{1.00,0.95,0.95} 3.93 &\cellcolor[rgb]{0.24,0.24,1.00} 5.24 &\cellcolor[rgb]{1.00,0.09,0.09} \fbox{\bf 1.72} \\
			&RFE& \cellcolor[rgb]{0.58,0.58,1.00} 4.67 &\cellcolor[rgb]{1.00,0.74,0.74} 3.71 &\cellcolor[rgb]{0.70,0.70,1.00} 4.53 &\cellcolor[rgb]{0.71,0.71,1.00} 4.53 &\cellcolor[rgb]{1.00,0.81,0.81} 3.79 &\cellcolor[rgb]{0.39,0.39,1.00} 4.84 &\cellcolor[rgb]{1.00,0.20,0.20} \fbox{\bf 1.93} \\
			&SU& \cellcolor[rgb]{0.93,0.93,1.00} 4.09 &\cellcolor[rgb]{1.00,0.82,0.82} 3.83 &\cellcolor[rgb]{0.46,0.46,1.00} 4.79 &\cellcolor[rgb]{0.99,0.99,1.00} 4.01 &\cellcolor[rgb]{0.88,0.88,1.00} 4.19 &\cellcolor[rgb]{0.20,0.20,1.00} 5.31 &\cellcolor[rgb]{1.00,0.16,0.16} \fbox{\bf 1.79} \\
			\cmidrule{1-9}
			\multirow{5}{*}{NB}
			&RF& \cellcolor[rgb]{0.87,0.87,1.00} 4.20 &\cellcolor[rgb]{0.75,0.75,1.00} 4.46 &\cellcolor[rgb]{1.00,0.66,0.66} 3.56 &\cellcolor[rgb]{0.16,0.16,1.00} 5.35 &\cellcolor[rgb]{0.81,0.81,1.00} 4.32 &\cellcolor[rgb]{1.00,0.59,0.59} 3.44 &\cellcolor[rgb]{1.00,0.30,0.30} \fbox{\bf 2.69} \\
			&RelF& \cellcolor[rgb]{0.69,0.69,1.00} 4.54 &\cellcolor[rgb]{0.97,0.97,1.00} 4.04 &\cellcolor[rgb]{1.00,0.72,0.72} \fbox{3.70} &\cellcolor[rgb]{0.65,0.65,1.00} 4.60 &\cellcolor[rgb]{1.00,0.77,0.77} \fbox{3.74} &\cellcolor[rgb]{1.00,0.61,0.61} \fbox{\bf 3.46} &\cellcolor[rgb]{1.00,0.94,0.94} \fbox{3.92} \\
			&SVM& \cellcolor[rgb]{0.78,0.78,1.00} 4.42 &\cellcolor[rgb]{0.79,0.79,1.00} 4.37 &\cellcolor[rgb]{1.00,0.77,0.77} 3.75 &\cellcolor[rgb]{0.37,0.37,1.00} 4.86 &\cellcolor[rgb]{1.00,0.96,0.96} 3.94 &\cellcolor[rgb]{1.00,0.46,0.46} \fbox{\bf 3.27} &\cellcolor[rgb]{1.00,0.54,0.54} \fbox{3.39} \\
			&RFE& \cellcolor[rgb]{0.72,0.72,1.00} 4.51 &\cellcolor[rgb]{0.89,0.89,1.00} 4.17 &\cellcolor[rgb]{1.00,0.72,0.72} \fbox{3.69} &\cellcolor[rgb]{0.60,0.60,1.00} 4.65 &\cellcolor[rgb]{1.00,0.90,0.90} 3.88 &\cellcolor[rgb]{1.00,0.55,0.55} \fbox{\bf 3.39} &\cellcolor[rgb]{1.00,0.74,0.74} \fbox{3.71} \\
			&SU& \cellcolor[rgb]{0.73,0.73,1.00} 4.49 &\cellcolor[rgb]{0.90,0.90,1.00} 4.16 &\cellcolor[rgb]{1.00,0.67,0.67} \fbox{3.58} &\cellcolor[rgb]{0.57,0.57,1.00} 4.70 &\cellcolor[rgb]{1.00,0.99,0.99} 3.98 &\cellcolor[rgb]{1.00,0.60,0.60} \fbox{\bf 3.45} &\cellcolor[rgb]{1.00,0.71,0.71} \fbox{3.65} \\
			\cmidrule{1-9}
			\multirow{5}{*}{RF}
			&RF& \cellcolor[rgb]{0.64,0.64,1.00} 4.63 &\cellcolor[rgb]{0.49,0.49,1.00} 4.76 &\cellcolor[rgb]{0.87,0.87,1.00} 4.20 &\cellcolor[rgb]{0.22,0.22,1.00} 5.30 &\cellcolor[rgb]{1.00,0.80,0.80} 3.77 &\cellcolor[rgb]{1.00,0.35,0.35} 3.08 &\cellcolor[rgb]{1.00,0.27,0.27} \fbox{\bf 2.27} \\
			&RelF& \cellcolor[rgb]{0.47,0.47,1.00} 4.78 &\cellcolor[rgb]{0.61,0.61,1.00} 4.65 &\cellcolor[rgb]{0.88,0.88,1.00} 4.18 &\cellcolor[rgb]{0.74,0.74,1.00} 4.48 &\cellcolor[rgb]{1.00,0.54,0.54} 3.38 &\cellcolor[rgb]{1.00,0.33,0.33} \fbox{\bf 2.82} &\cellcolor[rgb]{1.00,0.75,0.75} 3.72 \\
			&SVM& \cellcolor[rgb]{0.68,0.68,1.00} 4.57 &\cellcolor[rgb]{0.56,0.56,1.00} 4.70 &\cellcolor[rgb]{1.00,0.97,0.97} 3.96 &\cellcolor[rgb]{0.54,0.54,1.00} 4.72 &\cellcolor[rgb]{1.00,0.76,0.76} 3.73 &\cellcolor[rgb]{1.00,0.43,0.43} 3.19 &\cellcolor[rgb]{1.00,0.38,0.38} \fbox{\bf 3.13} \\
			&RFE& \cellcolor[rgb]{0.43,0.43,1.00} 4.80 &\cellcolor[rgb]{0.41,0.41,1.00} 4.82 &\cellcolor[rgb]{0.91,0.91,1.00} 4.14 &\cellcolor[rgb]{0.65,0.65,1.00} 4.62 &\cellcolor[rgb]{1.00,0.58,0.58} 3.43 &\cellcolor[rgb]{1.00,0.31,0.31} \fbox{\bf 2.81} &\cellcolor[rgb]{1.00,0.53,0.53} \fbox{3.37} \\
			&SU& \cellcolor[rgb]{0.50,0.50,1.00} 4.76 &\cellcolor[rgb]{0.38,0.38,1.00} 4.85 &\cellcolor[rgb]{0.85,0.85,1.00} 4.23 &\cellcolor[rgb]{0.63,0.63,1.00} 4.63 &\cellcolor[rgb]{1.00,0.47,0.47} 3.29 &\cellcolor[rgb]{1.00,0.34,0.34} \fbox{\bf 2.92} &\cellcolor[rgb]{1.00,0.50,0.50} 3.31 \\
			\cmidrule{1-9}
			\multirow{5}{*}{SVMG}
			&RF& \cellcolor[rgb]{0.61,0.61,1.00} 4.64 &\cellcolor[rgb]{0.68,0.68,1.00} 4.54 &\cellcolor[rgb]{1.00,0.64,0.64} 3.49 &\cellcolor[rgb]{0.67,0.67,1.00} 4.59 &\cellcolor[rgb]{0.94,0.94,1.00} 4.08 &\cellcolor[rgb]{1.00,0.65,0.65} 3.50 &\cellcolor[rgb]{1.00,0.39,0.39} \fbox{\bf 3.15} \\
			&RelF& \cellcolor[rgb]{0.55,0.55,1.00} 4.72 &\cellcolor[rgb]{0.84,0.84,1.00} 4.25 &\cellcolor[rgb]{1.00,0.71,0.71} 3.65 &\cellcolor[rgb]{0.80,0.80,1.00} 4.33 &\cellcolor[rgb]{1.00,0.70,0.70} 3.64 &\cellcolor[rgb]{1.00,0.42,0.42} \fbox{\bf 3.19} &\cellcolor[rgb]{0.85,0.85,1.00} 4.22 \\
			&SVM& \cellcolor[rgb]{0.36,0.36,1.00} 4.91 &\cellcolor[rgb]{0.81,0.81,1.00} 4.32 &\cellcolor[rgb]{1.00,0.51,0.51} \fbox{3.33} &\cellcolor[rgb]{0.72,0.72,1.00} 4.51 &\cellcolor[rgb]{1.00,0.84,0.84} 3.85 &\cellcolor[rgb]{1.00,0.40,0.40} \fbox{\bf 3.15} &\cellcolor[rgb]{1.00,0.93,0.93} 3.92 \\
			&RFE& \cellcolor[rgb]{0.54,0.54,1.00} 4.73 &\cellcolor[rgb]{0.82,0.82,1.00} 4.29 &\cellcolor[rgb]{1.00,0.69,0.69} 3.63 &\cellcolor[rgb]{0.75,0.75,1.00} 4.46 &\cellcolor[rgb]{1.00,0.68,0.68} 3.59 &\cellcolor[rgb]{1.00,0.52,0.52} \fbox{\bf 3.33} &\cellcolor[rgb]{1.00,0.98,0.98} \fbox{3.97} \\
			&SU& \cellcolor[rgb]{0.59,0.59,1.00} 4.65 &\cellcolor[rgb]{0.80,0.80,1.00} 4.33 &\cellcolor[rgb]{1.00,0.63,0.63} \fbox{\bf 3.48} &\cellcolor[rgb]{0.79,0.79,1.00} 4.35 &\cellcolor[rgb]{1.00,0.83,0.83} 3.83 &\cellcolor[rgb]{1.00,0.63,0.63} \fbox{3.48} &\cellcolor[rgb]{1.00,0.88,0.88} \fbox{3.86} \\
			\cmidrule{1-9}
			\multirow{5}{*}{SVML}
			&RF& \cellcolor[rgb]{0.66,0.66,1.00} 4.60 &\cellcolor[rgb]{0.35,0.35,1.00} 4.93 &\cellcolor[rgb]{1.00,0.84,0.84} 3.83 &\cellcolor[rgb]{0.18,0.18,1.00} 5.33 &\cellcolor[rgb]{0.95,0.95,1.00} 4.08 &\cellcolor[rgb]{1.00,0.59,0.59} 3.44 &\cellcolor[rgb]{1.00,0.13,0.13} \fbox{\bf 1.78} \\
			&RelF& \cellcolor[rgb]{0.63,0.63,1.00} 4.63 &\cellcolor[rgb]{0.29,0.29,1.00} 5.19 &\cellcolor[rgb]{1.00,0.85,0.85} 3.85 &\cellcolor[rgb]{0.59,0.59,1.00} 4.66 &\cellcolor[rgb]{1.00,0.78,0.78} 3.75 &\cellcolor[rgb]{1.00,0.65,0.65} 3.52 &\cellcolor[rgb]{1.00,0.29,0.29} \fbox{\bf 2.40} \\
			&SVM& \cellcolor[rgb]{0.51,0.51,1.00} 4.75 &\cellcolor[rgb]{0.33,0.33,1.00} 4.94 &\cellcolor[rgb]{0.93,0.93,1.00} 4.08 &\cellcolor[rgb]{0.30,0.30,1.00} 4.98 &\cellcolor[rgb]{1.00,0.96,0.96} 3.94 &\cellcolor[rgb]{1.00,0.45,0.45} 3.27 &\cellcolor[rgb]{1.00,0.22,0.22} \fbox{\bf 2.04} \\
			&RFE& \cellcolor[rgb]{0.53,0.53,1.00} 4.74 &\cellcolor[rgb]{0.27,0.27,1.00} 5.21 &\cellcolor[rgb]{1.00,0.91,0.91} 3.89 &\cellcolor[rgb]{0.34,0.34,1.00} 4.93 &\cellcolor[rgb]{1.00,0.87,0.87} 3.86 &\cellcolor[rgb]{1.00,0.36,0.36} 3.12 &\cellcolor[rgb]{1.00,0.26,0.26} \fbox{\bf 2.25} \\
			&SU& \cellcolor[rgb]{0.52,0.52,1.00} 4.74 &\cellcolor[rgb]{0.26,0.26,1.00} 5.23 &\cellcolor[rgb]{0.98,0.98,1.00} 4.03 &\cellcolor[rgb]{0.42,0.42,1.00} 4.81 &\cellcolor[rgb]{1.00,0.75,0.75} 3.72 &\cellcolor[rgb]{1.00,0.49,0.49} 3.30 &\cellcolor[rgb]{1.00,0.24,0.24} \fbox{\bf 2.17} \\ 
			\cmidrule{1-9}
		\end{tabular}
		
		\medskip
		\noindent
		Notes: C: Classifier, R: Ranker, RelF: ReliefF, RFE: SVM-RFE, SU: Symmetrical Uncertainty, RF: Random Forest (used as a classifier and as a ranker).
	}
\end{table}

The Friedman test was applied six times for each row of the table. We sorted the selectors from best to worst based on their average rank. Then we checked for difference between selectors at places 1 and 2, then between selectors at places 1, 2, and 3, then between 1, 2, 3, and 4, and so on. 
Take for example, the second row of the table where the classifier is 1NN and the Ranker is ReliefF. The test did not find a difference between the best selector (Top20) and the second best (RND20) significant at level 0.05. This is why they are both shown in boxes in Table~\ref{tab:res}. When we add the third-ranked selector however (Top10), the $p$-value is below the significance level. Obviously, once discovered, the difference is propagated down the list.

%

The results in Table~\ref{tab:res} give support to our cautionary tale. We observe that the selector's merit depends strongly on the {\em classifier}. The importance of the pair (feature-selector, classifier) has been often overlooked. Many studies propose new criteria for filter selection methods but it is not clear how those criteria will translate into classification accuracy using all returned features together. Striving for a consensus of classifiers, we may miss an extremely valuable subset of features which gives perfect results with just one classifier model. In reality, pairing a classifier with the feature set leads to dramatic differences. The two alternatives are: return $S$ which has moderate performance across many classifiers versus return $S$ which has excellent performance with one specific classifier.  Our example in Figure~\ref{fig:classifier-dependent} demonstrates this effect on a contrived example but Table~\ref{tab:res} gives the real-data perspective. With LDC and SVML, using ALL features is the winning strategy. In other words, any feature selection makes the result worse. For 1NN, NB and SVMG, the Top20 selector is the winner followed by RND20.  Interestingly, the best selector for 1NN, NB and SVMG is the {\em worst} selector for LDC, followed closely by RND20, once again highlighting the importance of classifier choice. 


There is little consistency among the ranking of the selectors across different classifiers. On average, ALL seems to dominate over the other selectors. On the other hand, with small exceptions, the behaviour of the selectors seems consistent over the ranking methods. This can be seen by the similar colours of the columns within the block of each classifier. For example, Top3 is worse than Top20 for 1NN, NB, RF, SVMG and SVML for any ranking method. This reinforces our message that developing new feature ranking algorithms may not be the best way forward to solving the wide data problem. Choosing Classifier, Ranker and Selector together holds a lot more promise.

\begin{figure}[h]
	\centering
	\includegraphics[width=0.6\linewidth]{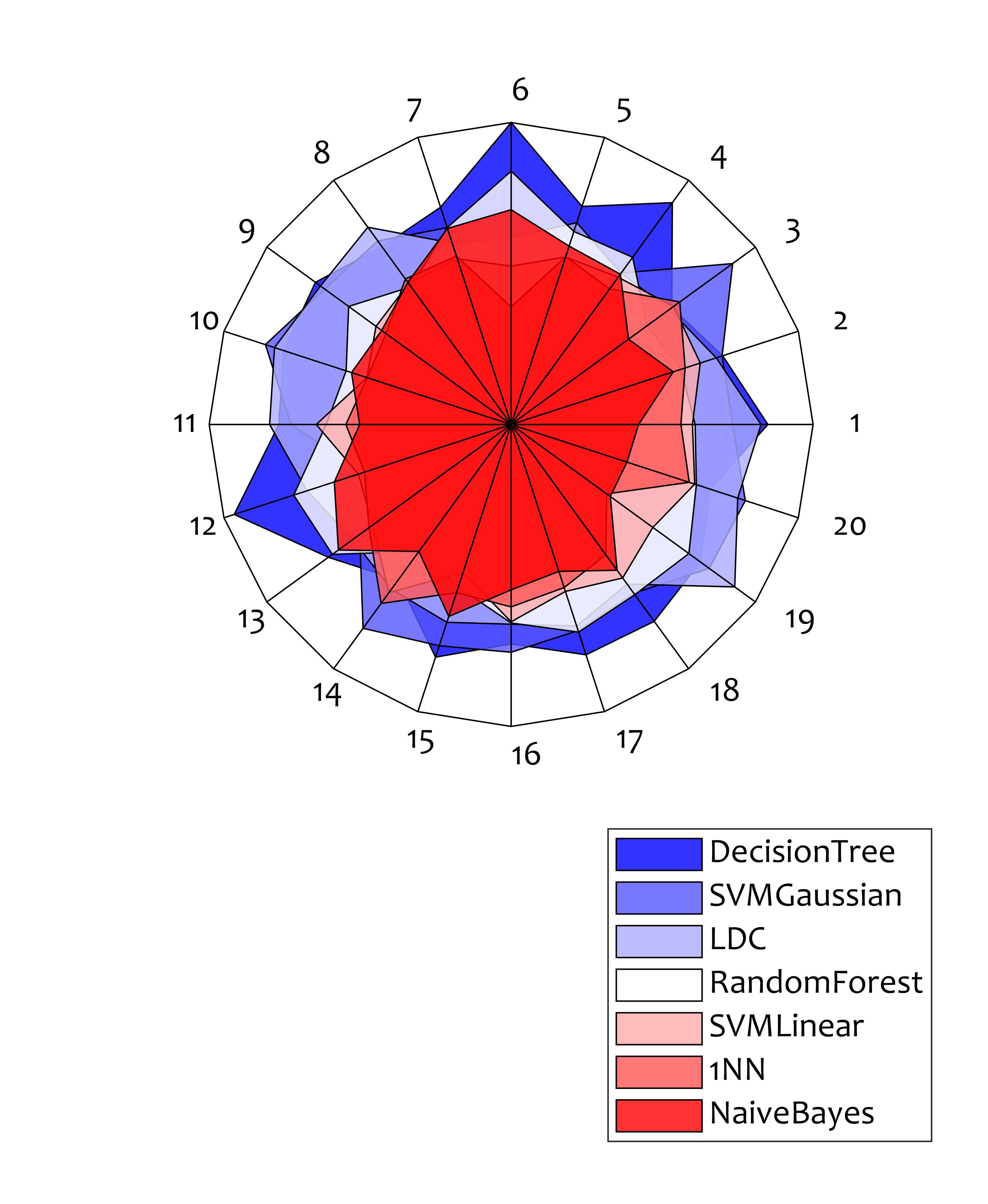}
	\caption{Glyph plot of the ranks of the seven classifiers for the 20 datasets.}
	\label{fig:g1}
\end{figure}

We chose glyph plots to visualise the rest of the results. The spokes on each plot correspond to the 20 datasets ordered as in Table~\ref{tab:data}. Figure~\ref{fig:g1} shows the ranks of the seven classifiers for the 20 datasets. The ranks are averaged across runs, rankers and selectors. The better classifier in this experiment was NB, and the worst was LDC.

\begin{figure}[h]
	\centering
	\includegraphics[width=0.6\linewidth]{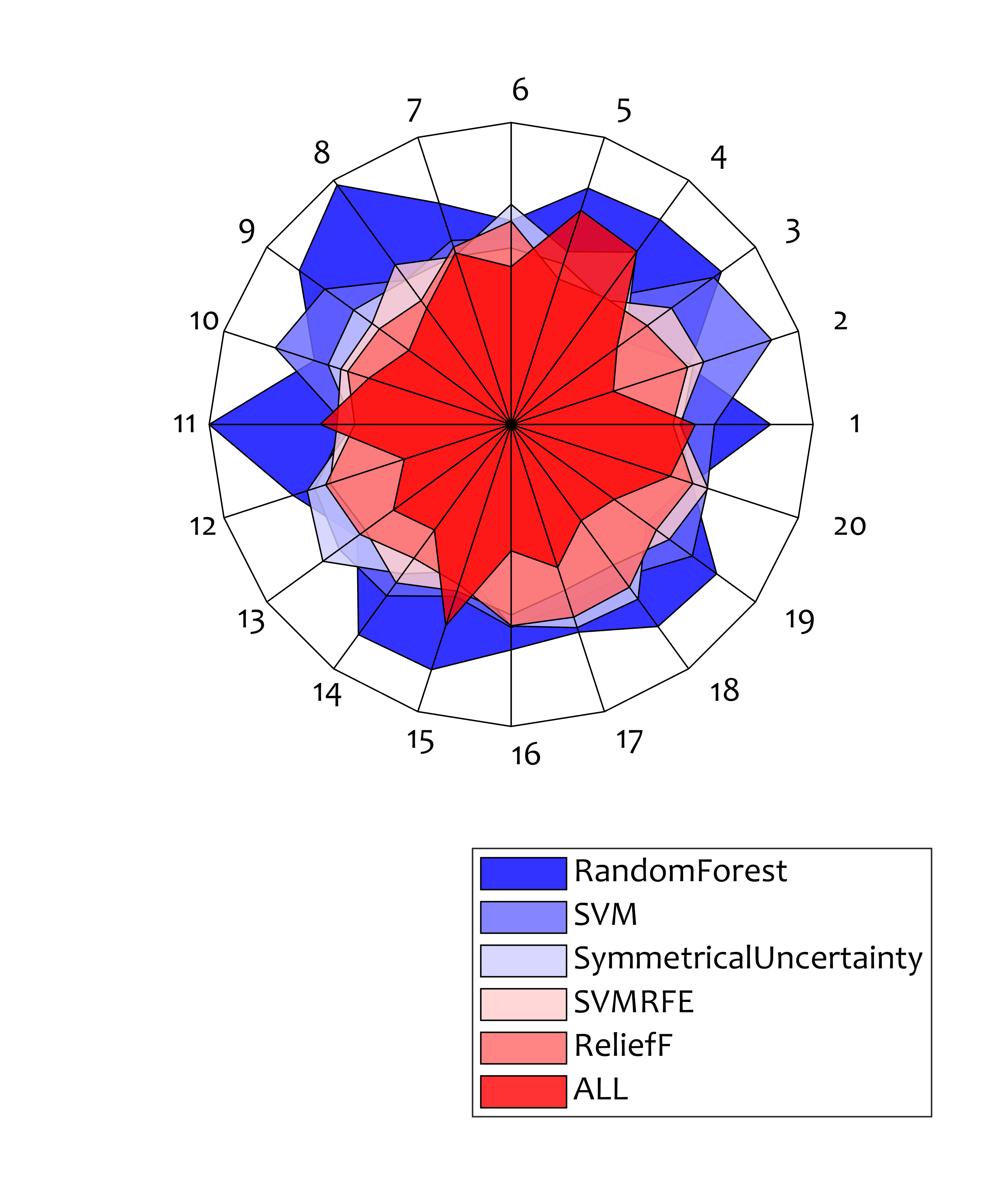}
	\caption{Glyph plot of the ranks of the five rankers for the 20 datasets. (Smaller ranks are better.)}
	\label{fig:g2}
\end{figure}

Figure~\ref{fig:g2} presents a similar view but this time the average is across runs, classifiers and selectors. Thus, the glyph plot shows the relationship between the different ranker methods. The legend is arranged from the worst method (the one with the largest area) to the best. Although we nominated ALL to be a selector, we can treat it as a ranker as well. Thus we included it in the comparison. According to the plot, the best strategy is ALL, followed by ReliefF and SVM-RFE. 

\begin{figure}[h]
	\centering
	\includegraphics[width=0.6\linewidth]{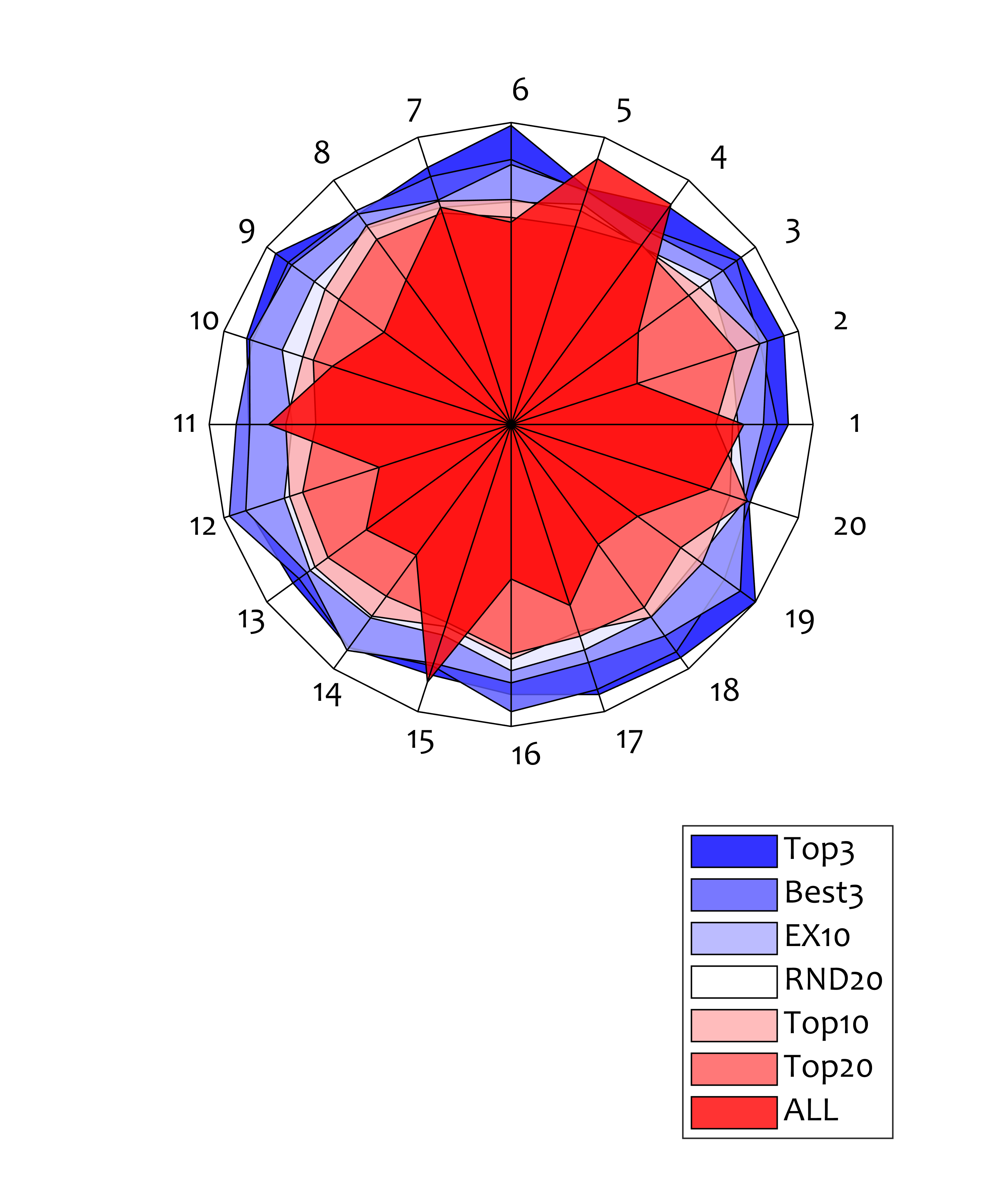}
	\caption{Glyph plot of the ranks of the selectors for the 20 datasets.}
	\label{fig:g3}
\end{figure}

Figure~\ref{fig:g3} shows the glyph plot for the selectors. Again, ALL wins, followed by Top20 and Top10.

Finally, for each data set, we arranged the 217 possible combinations of Classifier, Ranker and Selector as a matrix with 217 rows and 10 columns. Each column was the result of a run. We applied the Friedman test again to determine whether there is a statistically significant difference between the combinations. The top 35 combinations are shown in Table~\ref{taballranks}. Even with this small number of runs, the test clearly separated two of the combinations: both with ALL features, and both representing a linear classification rule (SVML and LDC)!
\begin{table}[p]
	\caption{Averaged ranks for the top 35 combinations of classifier, ranker and selector. The results are sorted from best to worst. The double horizontal line separates combinations which are significantly better than the rest.}
	\label{taballranks}
	\centering
	\bigskip
	\begin{tabular}{@{}rclll@{}}
		\toprule
		Position&Rank&Classifier&Ranker&Selector\\
		\midrule
		1 & 11.075 & SVML & -- & ALL\\
		2 & 14.000 & LDC & -- & ALL\\
		\hline\hline
		3 & 27.350 & SVML & RFE & Top20\\
		4 & 31.475 & SVML & RelF & Top20\\
		5 & 31.925 & NB & RelF & Top20\\
		6 & 32.275 & 1NN & SU & Top20\\
		7 & 33.300 & RF & RelF & Top20\\
		8 & 35.525 & 1NN & RelF & Top20\\
		9 & 35.875 & SVML & SU & Top20\\
		10 & 38.800 & 1NN & RFE & Top20\\
		11 & 39.300 & 1NN & RelF & RND20\\
		12 & 40.725 & NB & RelF & RND20\\
		13 & 42.100 & SVML & RelF & RND20\\
		14 & 42.600 & 1NN & SU & RND20\\
		15 & 43.175 & RF & RFE & Top20\\
		16 & 43.975 & NB & RelF & Top10\\
		17 & 44.975 & NB & RFE & Top20\\
		18 & 45.125 & SVML & RelF & Top10\\
		19 & 47.675 & 1NN & RelF & Top10\\
		20 & 48.125 & 1NN & RFE & RND20\\
		21 & 49.275 & SVML & SVM & Top20\\
		22 & 50.275 & RF & RelF & Top10\\
		23 & 50.300 & SVML & RFE & RND20\\
		24 & 51.725 & SVML & RFE & Top10\\
		25 & 53.025 & 1NN & SU & Top10\\
		26 & 53.100 & NB & SU & Top20\\
		27 & 53.175 & NB & RFE & RND20\\
		28 & 54.150 & RF & SU & Top20\\
		29 & 55.775 & NB & SU & RND20\\
		30 & 56.700 & 1NN & SVM & Top20\\
		31 & 57.875 & NB & RFE & Top10\\
		32 & 58.000 & 1NN & RFE & Top10\\
		33 & 58.000 & 1NN & SU & EX10\\
		34 & 58.250 & 1NN & RelF & EX10\\
		35 & 58.875 & 1NN & RFE & EX10\\
		\bottomrule
	\end{tabular}
\end{table}

The results from this experiment lead to the following overall recommendation. For very wide data sets, even for moderately-wide ones, it may be advisable to refrain from feature selection altogether. If we are prepared to accept a possible sacrifice of classification accuracy and go ahead with the feature selection, it is important to choose carefully the classifier model in conjunction with the selector and the ranking method. The combination of classifier and selector seems to be the more important factor than the ranking method. This is a curious result because a large proportion of the literature is devoted to perfecting the ranking (and selection) algorithms.


\section{Conclusion}
\label{con}

This paper is a warning against feature selection from datasets with very low sample size and a large number of original features, termed `wide' datasets. We describe the problem of estimating classification error as a criterion for feature selection and the accompanying caveats. A theoretical argument on the task of choosing one of two features amplifies the concern about the necessary sample size for a correct decision. It turns out that even for this simple problem, we may need hundreds of instances for a standard level of certainty in the right choice. An experiment with 20 real datasets sets apart our study from previous studies of similar nature, where the claims are usually illustrated by synthetic data with known (and well-behaved) distributions. 


The misconception that practitioners often rely on is that cross-validation will prevent overfitting in all circumstances. Valid as this may be for standard datasets, for wide datasets, cross-validation estimates offer only a marginal remedy. We have used the leave-one-out (LOO) estimate in all our experiments and examples, but the true error (estimated from the left-aside large portion of the data) did not correlate well with the estimates. This is evidenced by the fact that the smallest LOO error (guaranteed by the exhaustive search EX10) did not lead to the best testing error. This point was also illustrated by the example in Section~\ref{rel} (Figure~\ref{illu}).

Our experimental results revealed that considering the classifier that should be subsequently used with the selected features is paramount. The choice of classifier may determine whether we should attempt feature selection at all. In our experiment, the two linear classifiers performed best with ALL features, discouraging any of the feature selection alternatives. 
In all cases, the computationally intense selectors were found significantly inferior to the best selectors. This comes as no surprise, and is due to `overusing' the data, which is the major deficiency of wide datasets. 

But all is not lost! ALL (no selection) is not the best method for all datasets as seen in Figures~\ref{fig:g2} and \ref{fig:g3}. In trying to answer the general question, we have bypassed another crucial one: What is best for {\em my} wide dataset? This is a difficult question in its own right, and deserves a separate study. 

Admittedly, our experiment exaggerated the problem by creating an unrealistic scenario with only 10 samples per class in the sample. The problems will naturally be mitigated with increase of the sample size. An interesting future research line would be to evaluate the severity of the issues as a function of the sample size $N$. This brings the question of whether stability of the selected subsets may be an indicator of whether feature selection should be attempted and which selectors and classifiers should be preferred.

\section*{Acknowledgements}
This work was done under project RPG-2015-188 funded by The Leverhulme Trust, UK, project TIN2015-67534-P (MINECO/FEDER, UE) of the \emph{Ministerio de Econom\'ia y Competitividad} of the Spanish Government, and project BU085P17 (JCyL/FEDER, UE) of the \emph{Junta de Castilla y Le\'on} (both projects cofinanced through European Union FEDER funds).
The third author was supported by the Mobility Grant CAS19/00100 from the \emph{Ministerio de Ciencia, Innovaci\'on y Universidades} of the Spanish Government.



\end{document}